\documentclass[letterpaper, 10 pt, conference]{ieeeconf}  

\IEEEoverridecommandlockouts                              

\usepackage{cite}
\usepackage{graphics} 
\usepackage{epsfig} 
\usepackage{mathptmx} 
\usepackage{times} 
\usepackage{amsmath} 
\usepackage{amssymb}  

\usepackage{hyperref}

\title{\LARGE \bf
HANDO: Hierarchical Autonomous Navigation and Dexterous Omni-loco-manipulation

}

\author{
Jingyuan Sun$^{* \ddagger}$, Chaoran Wang$^{* \dagger}$, Mingyu Zhang$^{* \dagger}$, Cui Miao$^{* \dagger}$, Hongyu Ji$^{\dagger}$, Zihan Qu$^{\dagger}$, Han Sun$^{\dagger}$\\
Bing Wang$^{\ddagger}$,  Qingyi Si$^{\ddagger}$ 
\thanks {$^{*}$These authors contributed equally to this work.}
\thanks{\(^{\dagger}\)This work was performed during an internship at Huawei.}
\thanks{\(^{\ddagger}\)Corresponding author: Jingyuan Sun, Bing Wang and Qingyi Si are with Shanghai Huawei Technologies Co., Ltd., Shanghai 201799, China (e-mail: sunjingyuan1@huawei.com, wangbing183@huawei.com, siqingyi@huawei.com).}
}

\begin{document}

\maketitle
\thispagestyle{empty}
\pagestyle{empty}

\begin{abstract}

Seamless loco-manipulation in unstructured environments requires robots to leverage autonomous exploration alongside whole-body control for physical interaction.
In this work, we introduce HANDO (Hierarchical Autonomous Navigation and Dexterous Omni-loco-manipulation), a two-layer framework designed for legged robots equipped with manipulators to perform human-centered mobile manipulation tasks.
The first layer utilizes a goal-conditioned autonomous exploration policy to guide the robot to semantically specified targets, such as a black office chair in a dynamic environment. The second layer employs a unified whole-body loco-manipulation policy to coordinate the arm and legs for precise interaction tasks—for example, handing a drink to a person seated on the chair. 
We have conducted an initial deployment of the navigation module, and will continue to pursue finer-grained deployment of whole-body loco-manipulation.
The video can be found at (\url{https://youtu.be/YD0qx3vRsfc}).

\end{abstract}

\section{INTRODUCTION}

Last-mile delivery has emerged as a critical application for service robots, in which these robotic systems must not only traverse complex environments but also physically interact with humans. 
Conventional delivery approaches often rely on pre-built maps and precise localization 
\cite{hoshi2022graph}.
While effective in structured settings, these methods incur high costs for map construction and limit scalability to customized or dynamic environments. 
In parallel, last-mile delivery tasks inherently require dexterous whole-body interaction-for instance, grasping a takeaway bag and handling it to a seated person, which highlight the need to unify robust locomotion with manipulation.

Dexterous omni-loco-manipulation
with onboard manipulators represents a natural platform for last-mile delivery. 
Quadrupedal robots provide agile locomotion and the ability to traverse uneven terrain~\cite{miki2022learning}, 
while the arm introduces manipulation capabilities such as grasping, carrying and handover. Compared to standalone manipulators, quaduped-with-arm systems offer superior mobility and exploration; compared to legged robots alone, they extend functionality by enabling rich physical interactions with humans and objects. This synergy positions legged mobile manipulators as a powerful embodiment for complex delivery scenarios, combining the flexibility of locomotion with dexterity of manipulations.

On the navigation side, most delivery strategies remain map-based, relying on prior environmental models~\cite{hoshi2022graph}, which struggle in frequently changing or rapidly deployed settings. To address these limitations, recent works such as Navila~\cite{cheng2024navila}, FRTree planner~\cite{li2025frtree}, and  Unigoal~\cite{yin2025unigoal}
investigate map-free navigation, enabling robots to reason about and traverse unknown spaces directly from sensory observations. While these approaches advance the feasibility of zero-cost deployment, significant opportunities remain to further improve autonomous exploration and adaptability.

On the manipulation side, recent work emphasizes the importance of whole-body coordination. Methods such as UMI-on-Legs\cite{ha2024umi} and MLM\cite{liu2025mlm} employ reinforcement learning (RL) and trajectory-driven policies to generate coordinated arm-leg behaviors guided by end-effector trajectories. Diffusion-based policies\cite{chi2023diffusion} further enhance flexibility by learning trajectory distributions that are generalized among tasks. Such approaches enable quadruped to perform loco-manipulation tasks ranging from pushing to handover. 

Compared to simulation, 
real-world deployment poses significant hurdles. Perception gaps, terrain variability, and hardware constraints limit zero-shot transfer. 
Moreover, delivery entails integrated autonomy that combines exploration and execution: posing the challenge of enabling robots to autonomously navigate unknown spaces, 
and then execute dexterous manipulator actions for human interaction. Bridging these gaps demands a hierarchical yet integrated framework that unifies map-free navigation with whole-body loco-manipulation in a deployable system.

\begin{figure*}[!t]
		\centering
		\includegraphics
        [width=0.8\textwidth]
        {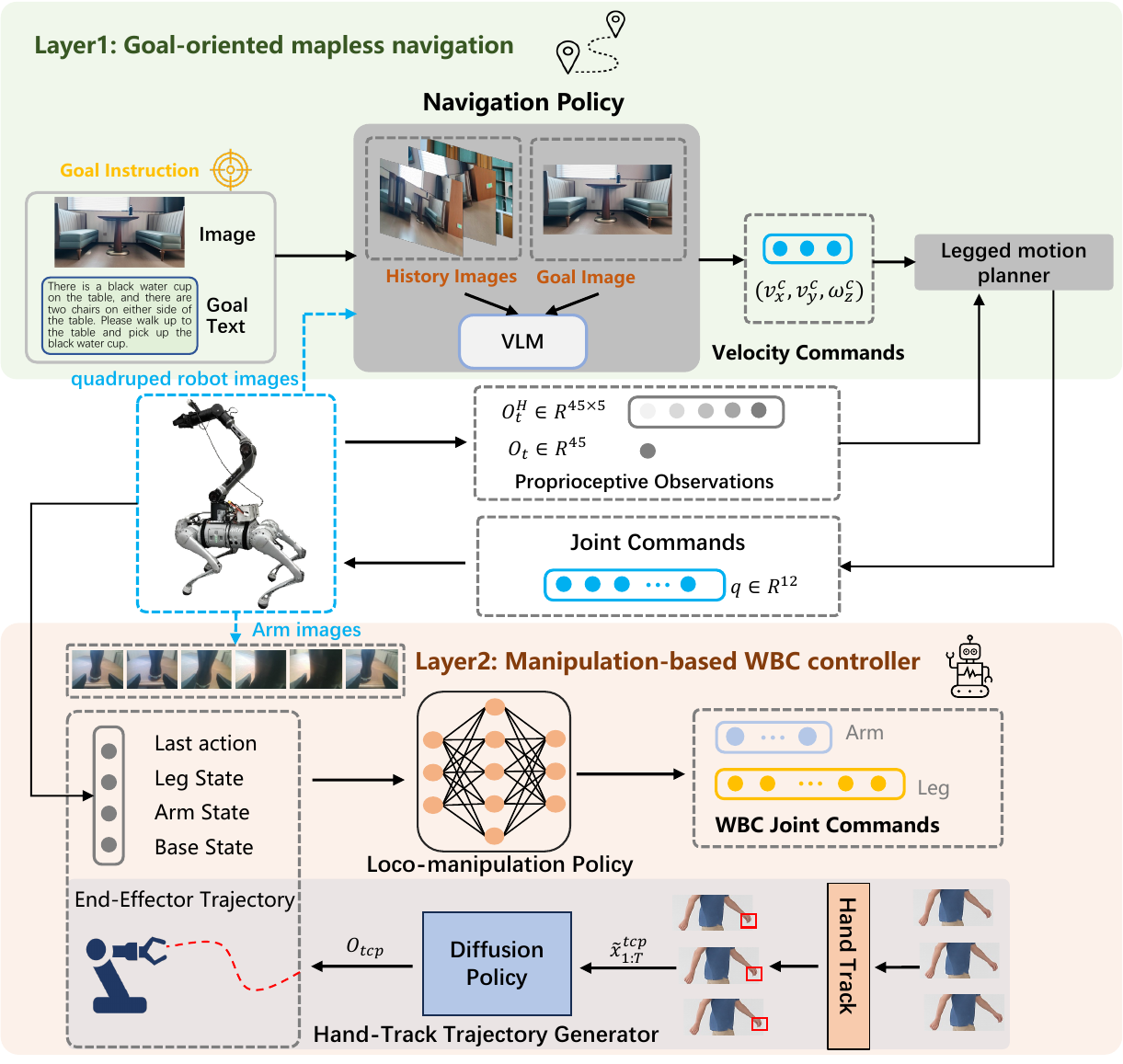}
		\caption{Overview of HANDO. The two-layer framework couples mapless navigation (Layer 1) with whole-body loco-manipulation (Layer 2), where navigation outputs velocity/joint commands and manipulation uses hand-track with diffusion policy to generate coordinated grasping and handover.}
		\label{fig:HANDO sys}
	\end{figure*}

To address the above issues,
we developed a two-layer framework $\bf{HANDO}$ for last-mile delivery that enabled seamless coordination between navigation and loco-manipulation controls. Our contributions are summarized as follows:
\begin{itemize}
\item 
We propose a novel map-free navigation module, which employs a vision-language model for cross-scene reasoning and graph matching to drive a three-stage exploration strategy, thereby enabling zero-cost navigation without pre-built maps.

\item 
We propose a loco-manipulation policy that fuses quadruped locomotion and arm control, guided by end-effector trajectories, to achieve whole-body interaction behaviors such as grasping and handover.

\item 
We integrate and validate the system on a real quadruped-with-arm platform (see Fig. \ref{fig:HANDO sys}), demonstrating end-to-end last-mile delivery that combines semantic navigation and whole-body interaction in human centered scenarios.
\end{itemize}

\section{Related Work}
\subsection{Autonomous Navigation}
Navigation for delivery and service robots has traditionally relied on map-based methods such as SLAM and graph-based planning~\cite{grisetti2011tutorial,hoshi2022graph}. While effective in static and structured settings, these approaches demand costly pre-mapping and struggle in dynamic or customized environments. Recent advances have sought to eliminate explicit maps by directly grounding navigation in perceptual inputs. For example, UniGoal~\cite{yin2025unigoal} introduces a goal-conditioned exploration framework that leverages language and vision cues to guide long-horizon navigation without map construction. Similarly, vision-language navigation framework such as NaviLa~\cite{cheng2024navila} demonstrate the potential of grounding navigation in semantic goals specified by natural language. These approaches significantly reduce deployment costs and improve adaptability, but they remain limited when coupled with manipulation, which is essential for last-mile delivery.

\subsection{End-to-end Imitation learning}
Imitation learning (IL) has achieved remarkable progress in robotic manipulation, directly mapping raw sensory inputs to robot actions from expert demonstrations.
Architecture such as ACT~\cite{george2023one} employ transformer backbones with image encoders to capture variability in human data, while Diffusion Policy~\cite{chi2023diffusion} introduces generative diffusion processes to model multimodel action distributions. Extensions have also explored 3D perception: Rise~\cite{saunders2006rise} leverages sparse point cloud encoders for continuous control.
However, while these methods enrich sensory representations, they often neglect explicit object-pose reasoning and introduce redundant information, leading to increased computational complexity and reduced generalization. Applying IL in mobile manipulation thus requires better integration of spatial reasoning with efficient whole-body control.

\subsection{Loco-Manipulation}
To endow legged robots with manipulation capabilities, researchers have integrated quadruped mobility with robotic arms. Early approaches employed optimization-based frameworks for footstep planning and whole-body trajectory generation~\cite{ha2024umi,liu2025mlm}, but these require task-specific design and incur high computational costs. More recent work has shifted toward reinforcement learning (RL) to achieve end-to-end control across multiple loco-manipulation tasks. For instance, \cite{gold2019external} modeled arm motions as external torques for locomotion compensation, while~\cite{fu2023deep} proposed Regularized Online Adaptation for sim-to-real transfer of whole-body control. However, these methods often restrict manipulation to arm-based coordinates, making them susceptible to body motion disturbances. Recent frameworks such as MLM~\cite{liu2025mlm} combine trajectory libraries and diffusion-policy-based reasoning to enable vision-driven, task-space loco-manipulation. Yet, challenges reamin in balancing performance across taks and integrating automatic task execution, especially under dynamic human-centered delivery scenarios.

\section{Goal-oriented mapless navigation}
\label{sec: navigation}

\subsection{Navigation Policy}
To achieve adaptive navigation without pre-built maps, we design a goal-oriented mapless navigation policy inspired by UniGoal~\cite{yin2025unigoal}. The central idea is to guide the quadruped robot toward a semantic goal without relying on explicit maps or localization, but instead leveraging \emph{goal-image matching} and reasoning from a vision-language model (VLM). Our method follows a three-stage exploration process as shown in Fig. \ref{fig:HANDO sys}.

\textbf{Stage 1: Initial Exploration.} The robot starts by scanning the environment using onboard RGB-D sensing and builds an incremental scene graph. When the matching score with the goal graph is below a threshold $\sigma_1$, i.e.,
\begin{equation}
s_t < \sigma_1,
\end{equation}
the system decomposes the semantic goal graph $G_g$ into sub-goals and employs frontier-based exploration, guided by a VLM, to cover unexplored areas.


\textbf{Stage 2: Coordinate Projection and Alignment.} Once partial matching is achieved, i.e.,
\begin{equation}
\sigma_1 \leq s_t < \sigma_2,
\end{equation}
the goal graph $G_g$ and current scene graph $G_t$ are aligned. 

\textbf{Stage 3: Goal Verification.} When the matching score exceeds $\sigma_2$, i.e.,
\begin{equation}
s_t \geq \sigma_2,
\end{equation}
the policy performs goal verification and scene graph correction to finalize navigation.

\textbf{Action Generation.} At Stage 2 and Stage 3, 
a VLM-based action decoder 
selects a discrete action $a_t \in \{\texttt{move forward}, \texttt{turn left}, \texttt{turn right}, \texttt{stop}\}$ by maximizing the expected improvement of the matching score.
These actions are mapped into continuous velocity commands: 
$\left(0.1~\text{ms}^{-1},\ \pi/12~\text{rad}~\text{s}^{-1},\ -\pi/12~\text{rad}~\text{s}^{-1},\ 0 \right)$
The continuous actions are then executed by the low-level legged locomotion controller. 
\section{Loco-Manipulation Policy}
\label{sec:loco-mani}

This part is a proposed design, and we are currently building a whole-body control simulation platform and collecting real-world data. In the future, the loco-manipulation module within the framework will be gradually implemented and refined. 

We address the challenge of last-mile delivery using a legged robot equipped with an onboard arm, which must perform integrated locomotion and manipulation to hand over or receive parcels from humans in unstructured environments. To improve human–robot motion compatibility, we propose a whole-body controller that is directly conditioned on real-time human hand trajectories. Our method has two modules: a \emph{Hand-Track Trajectory Generator} that converts human-hand observations into a smooth end-effector target trajectory, and a \emph{Whole-Body Loco-Manipulation Policy} outputs joint-space actions for legs and arm under a unified control policy.

\subsection{Hand-Track Trajectory Generator}
We detect the operator's hand, select keyframes by hand-speed troughs, and attach visual prompts to disambiguate spatial relations before VLM reasoning to produce coarse sub-goals~\cite{bakker1996robot}. 

Given the keyframes $\{I_{k}\}_{k=1}^{K}$ and their hand poses $\{\mathbf{h}_k\}$ in the camera frame, we compute a calibrated world-frame sequence $\tilde{\mathbf{x}}^{\text{tcp}}_{1:T}$ by retargeting hand positions/orientations to the robot gripper's tool-center-point (TCP) with a fixed transform ${}^{\text{tcp}}\mathbf{T}_{\text{hand}}$:
\begin{equation}
\mathbf{x}^{\text{tcp}}_t
= \mathrm{SE(3)}\!\left(\mathbf{T}_{\text{cam}\rightarrow \text{world}}\right)\!\cdot \mathrm{SE(3)}\!\left(\mathbf{h}_t\right)\!\cdot {}^{\text{tcp}}\mathbf{T}_{\text{hand}}.
\end{equation}

At deployment, the generator runs in two modes: \emph{online} from live hand tracking during human-robot handover, or \emph{offline} from pre-recorded human demonstration videos.

\subsection{Whole-Body Loco-Manipulation Policy}



We formulate the problem as a Partially Observable Markov Decision Process (POMDP) and learn a policy $\pi_\theta$ that outputs joint position offsets for both the legs and the arm. The policy is trained with PPO to maximize the expected discounted return:
$\mathbb{E}_{\tau(\pi_\theta)}\big[\sum_{t}\gamma^t r_t\big]$.

\textbf{State Space:}
The state for the policy includes the last action, leg state, arm state, base state and end-effector trajectory. Specifically, the leg state includes the joint positions $\mathbf{q}_{t}^{l} \in \mathbb{R}^{12}$, joint velocities $\mathbf{\dot{q}}_{t}^{l} \in \mathbb{R}^{12}$, the gravity vector $\mathbf{g}_{t} \in \mathbb{R}^{3}$ and the previous action $\mathbf{a}_{t-1}^{l} \in \mathbb{R}^{12}$. The arm state includes the joint positions $\mathbf{q}_{t}^{a} \in \mathbb{R}^{12}$, joint velocities $\mathbf{\dot{q}}_{t}^{a} \in \mathbb{R}^{12}$ and the previous action $\mathbf{a}_{t-1}^{a} \in \mathbb{R}^{6}$. 
The base state includes the body angular velocity  $~\mathbf{ \omega}_{t}^{b} \in \mathbb{R}^{3}$ and the body linear velocity $~\mathbf{v}_{t}^{b} \in \mathbb{R}^{3}$.
The end-effector trajectory is represented through a 3D position vector and a 6D rotation representation.

\textbf{Action Space:}
We use position PD control with target $\mathbf{q}^{*}_t=\mathbf{q}^{\text{default}}+\Delta \mathbf{q}_t$, consistent with the whole-body joint-space actuation.

\textbf{Rewards:}
The primary goal is TCP tracking in position and orientation:
\begin{equation}
r_{\text{track}}
= \exp\!\Big(-\tfrac{\lVert \mathbf{p}^{\text{tcp}}_t-\mathbf{p}^{\text{tar}}_t\rVert}{\sigma_p}\Big)\;
\cdot\;
\exp\!\Big(-\tfrac{\angle\!\big(\mathbf{R}^{\text{tcp}}_t (\mathbf{R}^{\text{tar}}_t)^\top\big)}{\sigma_o}\Big)
\end{equation}
We add regularizers for smoothness and hardware safety:
\begin{equation}
r_{\text{reg}}=-\lambda_\tau\lVert \boldsymbol{\tau}_t\rVert^2
-\lambda_{\Delta q}\lVert a_t-a_{t-1}\rVert^2
-\lambda_{\ddot q}\lVert \ddot{\mathbf{q}}_t\rVert^2,
\end{equation}
and optionally incorporate a locomotion style prior to stabilize gaits during manipulation \cite{liu2025mlm}. 
The total reward is $r_t=r_{\text{track}}+r_{\text{reg}}+r_{\text{style}}$.

\textbf{Policy architecture:}
We adopt a lightweight MLP actor-critic with a memory of recent proprioception. PPO is used for optimization. To improve sim-to-real, the standard domain randomization is used on PD gains, link masses, ground friction, and observation latency.





\section{Experiments and results}
\label{sec:exp&res}
\subsection{Experimental setup}
For hardware integration, we assembled a quadruped-manipulator platform consisting of a Unitree Go1 EDU robot and a AGILEX PIPER lightweight arm. The computation is provided by an NVIDIA RTX 4090 GPU, which enables multi-threaded control for real-time operation. Both the locomotion policy and the whole-body loco-manipulation policy run at 50Hz, ensuring repinsive execution of leg and arm actions. The generated joint trajectories are transmitted from the computer to the robot actuators via a wired Ethernet connection, supporting low-latency and reliable deployment in physical environments.

\subsection{Real-world Experiments}

The real-world evaluation was conducted in a café with an unstructured layout of irregularly arranged tables, chairs, and miscellaneous objects. In each trial, the robot started from a randomized initial position and was tasked with autonomously approaching a designated human recipient. The setting was partially observable: the robot had no prior knowledge of the target location and relied solely on visual input and semantic commands.

\begin{figure}[!t]
		\centering
		\includegraphics
        [width=88mm]
        {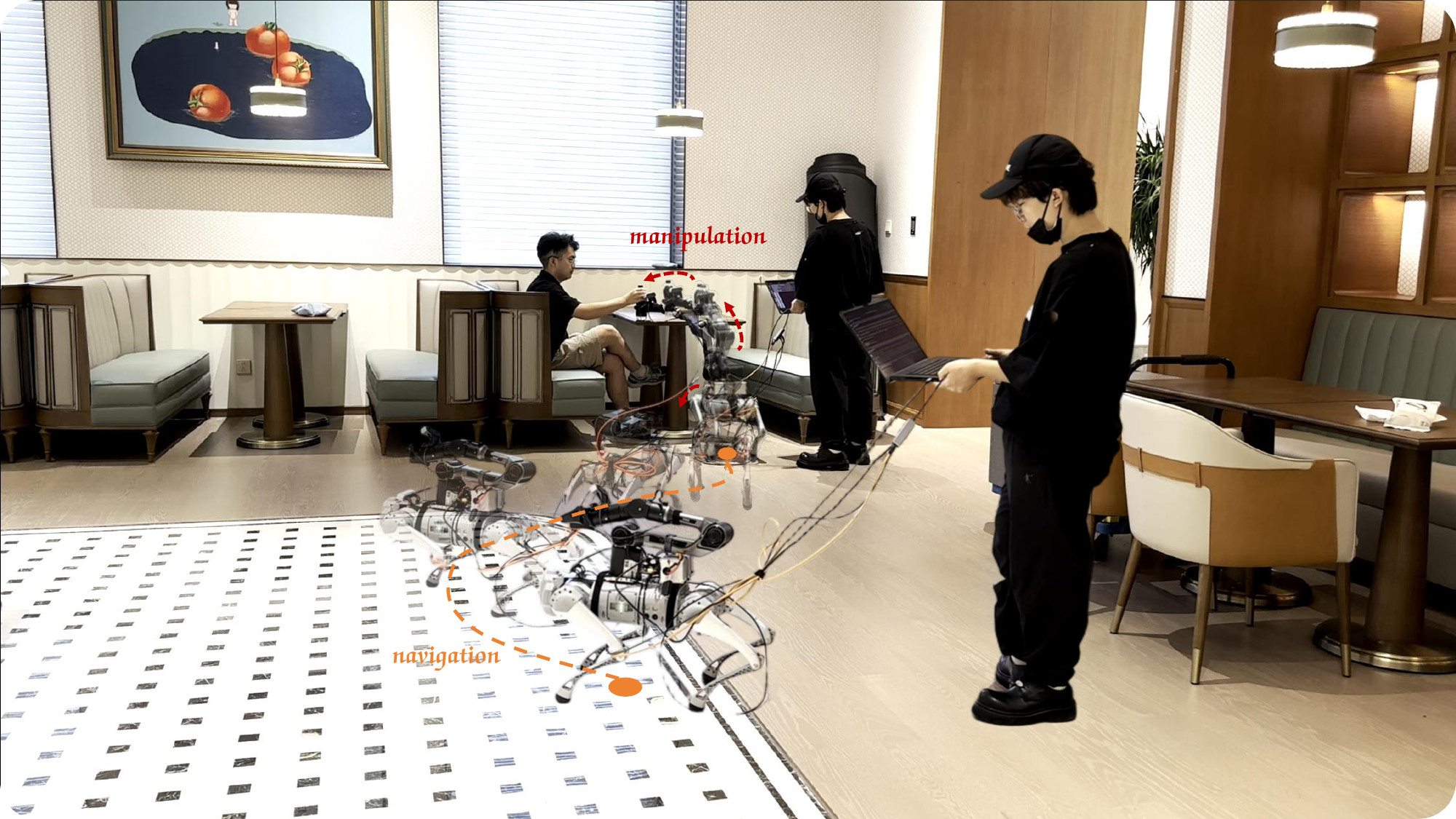}
	\caption{Snapshots of real-world experiments.The task
required the robot to deliver a beverage and handle to a seated human.}
		\label{fig: exp}
	\end{figure}

The goal-directed, mapless navigation layer consistently explored the environment and approached the target using only visual cues. Recorded base trajectories were smooth and continuous, indicating stable and robust navigation despite layout irregularities. Illustrative examples of the scene and the resulting navigation paths are provided in Fig.~\ref{fig: exp}.

\section{Conclusion and Future Work}
\label{sec:conclusion}

In this work, we present HANDO, a hierarchical framework for autonomous navigation and dexterous loco-manipulation, designed to enable legged mobile manipulators to perform complex last-mile delivery tasks in unstructured and human-populated environments. HANDO combines a high-level goal-conditioned exploration policy for map-free navigation with a unified whole-body policy for coordinating locomotion and manipulation, thereby bridging semantic task understanding with low-level physical control.

Beyond navigation, we will continue to elaborate on finer-grained training and deployment of whole-body loco-manipulation, with an emphasis on coordinated grasp-and-handover between the quadruped base and a manipulator.
In particular, we plan to integrate real-time hand tracking, enabling the robot to dynamically align its manipulator with the human hand during object transfer, thereby enhancing the safety, robustness, and naturalness of human–robot interaction in complex real-world settings.

\addtolength{\textheight}{-12cm}   


\bibliographystyle{bib/IEEEtran}
\bibliography{bib/sn-bibliography}

\end{document}